\documentclass[letterpaper]{article} 
\usepackage{aaai2026}  
\usepackage{times}  
\usepackage{helvet}  
\usepackage{courier}  
\usepackage[hyphens]{url}  
\usepackage{graphicx} 
\usepackage{natbib}  
\usepackage{caption} 
\usepackage{algorithm}
\usepackage{algorithmic}
\usepackage{pdfpages}

\usepackage{newfloat}
\usepackage{listings}
\DeclareCaptionStyle{ruled}{labelfont=normalfont,labelsep=colon,strut=off} 
\floatstyle{ruled}
\newfloat{listing}{tb}{lst}{}
\floatname{listing}{Listing}
\usepackage{amssymb}
\usepackage{multirow}
\usepackage{booktabs}
\usepackage{tabularx}
\usepackage[many]{tcolorbox}    	
\usepackage{cleveref}

\definecolor{main}{HTML}{5989cf}    
\definecolor{sub}{HTML}{cde4ff}     
\newtcolorbox{boxD}{
    colback = sub, 
    colframe = main, 
    boxrule = 0pt, 
    toprule = 3pt, 
    bottomrule = 3pt 
}

\title{Negotiating Risk Boundaries in AI for Policing Through \\ Mixed-Stakeholder Deliberation}
\author{
    Mackenzie Jorgensen\textsuperscript{\rm 1}\equalcontrib,
    Jo Reilly\textsuperscript{\rm 2}\equalcontrib,
    Alex Sutherland\textsuperscript{\rm 3},
    Miri Zilka\textsuperscript{\rm 2}\textsuperscript{\dag}
}
\affiliations{
    \textsuperscript{\rm 1}Northumbria University, UK\\
\textsuperscript{\rm 2}University of Cambridge, UK\\
\textsuperscript{\rm 3}University of Oxford, UK\\
\textsuperscript{\dag} mz477@cam.ac.uk

}

\usepackage{bibentry}

\begin{document}

\maketitle

\begin{abstract} 
\looseness=-1
AI tools are being increasingly adopted in policing in the UK and worldwide. Racial bias is a known and well-documented risk, yet representatives of affected communities are rarely included in decisions about AI adoption. We present results from a mixed-stakeholder deliberation workshop bringing together 30 community representatives, police officers, and academics to assess the risks of 13 AI use cases in policing, with an explicit focus on racial bias. We found that participants were broadly open to AI adoption, rejecting only three use cases outright -- most notably recidivism risk assessment, where objections targeted the premise rather than the implementation. Our analysis reveals that foregrounding racial equity did not narrow the deliberation. Instead, discussions gravitated toward a fundamental set of questions: does this tool actually work, will it deliver genuine benefit, and will that benefit extend to everyone? This integrated reasoning---reminiscent of the curb-cut effect in inclusive design---highlights
the benefit of incorporating the racial bias lens into the risk-benefit analysis of AI use cases from the outset. 
\end{abstract}



\section{Introduction}
As law enforcement agencies become increasingly stretched and resource-constrained, artificial intelligence (AI) is viewed as a key strategy for increasing efficiency in operations~\cite{NPCCAICovenant, AIplaybook2025}. Agencies have been experimenting with data-driven applications for over a decade — from predictive policing to facial recognition — and recent years have added significant political pressure towards large-scale adoption. Many concerns have arisen alongside this expansion. Among them, the persistence and further amplification of racial bias. The historical and ongoing disproportionality in the criminal justice system — where Black individuals are highly overrepresented in arrests, incarceration, and victimisation — makes racial bias a particularly urgent and hard-to-resolve challenge for AI in policing.

While the number of individual AI applications, even in the UK alone, is numerous~\cite{taka2025mapping}, AI use cases in policing fall into similar categories. Amongst the most popular and well-known are \textit{predictive policing} and \textit{recidivism risk assessment}, for both of which there is substantial academic literature documenting how they can perpetuate racial bias \citep{ensign2018runaway,zilka2023progression}. Responsible AI guidelines in policing have emerged in response, such as the AI Playbook for Policing~\citep{AIplaybook2025}, which advises forces to map which groups could be differentially affected and ensure evaluations capture race disproportionalities. However, they offer little guidance on \emph{how} evaluations should be conducted, and in practice, these are rarely performed or shared publicly. 

The people best placed to identify these risks \emph{early} are rarely consulted at the design stage. The teams tasked with addressing racial bias within policing are often siloed from those driving AI adoption, and the communities most affected rarely have a seat at the table. These decisions are typically driven by management and technical teams, and do not include those with direct knowledge of how racial disproportionality operates in practice. Earlier intervention, at the ideation stage, offers a more effective opportunity to flag approaches that could embed or exacerbate racial disproportionality before sunk costs make course-correction difficult.

We propose and demonstrate a method for negotiating acceptable risk limits within mixed-stakeholder groups, suitable for the conception stage of AI adoption. During a one-day in-person workshop, we convened 30 stakeholders spanning community representatives, police, and academia to consider 13 AI use cases in policing. Participants deliberated in mixed groups using a red, amber, and green risk framework (\Cref{tab:feedback_form}). Community representatives were not required to have technical knowledge; their contributions drew on expertise and lived experience of how racial disproportionality becomes embedded in the criminal justice system. Our aims were twofold: to establish which use cases participants found acceptable for deployment and under what conditions, and to understand the process by which they arrived at these judgments. Participants were broadly open to AI adoption, rejecting only three use cases outright, but were sceptical of AI's ability to drive meaningful improvements in policing. We found that foregrounding racial equity did not narrow the deliberation: instead, discussions focused on fundamental questions about whether a tool works, for whom, and under what conditions. We argue this is reminiscent of the curb-cut effect in inclusive design~\cite{blackwell2017curb,shneiderman2020bridging} — that designing with marginalised communities in mind surfaces better questions and can produce better outcomes for everyone.

\section{Background and Related Works}

\subsection{AI Use Cases in Policing}

There is substantial interest in AI use in law enforcement, in the UK and worldwide. What `counts' as AI varies, but below we give a brief background to a number of high-profile use cases that were considered in the workshop.  

\paragraph{Forecasting Crime Hotspots (``Predictive Policing'').}
Efforts to identify \textit{where} crime is likely to occur or stable hotspots to guide resource deployments have been undertaken from as early as the 1930s~\citep{shaw1942juvenile, braga2019hot}. This is usually based on the assumption that where crime has occurred repeatedly is where it is likely to happen in the future. However, realistically, accurate prediction of where \textit{and} \textit{when} crimes will occur is quite challenging~\citep{govuk2025ai}. 
The utility of ``real-time'' hotspot methods is questionable due to the finer the resolution, the variability, and the low number of recorded crimes. Indeed, evaluations of these prediction tools, such as PredPol$^{TM}$ have not shown evidence of crime prevention benefits~\citep{hunt2014shreveport,saunders2016predictions}. In addition, these tools raise strong concerns about racial inequity~\citep{lum2016predict,sankin2021crime}, even if they improve forecast accuracy~\citep{mohler2015randomized, lee2024effectiveness}. However, identifying \emph{stable} hotspots of violence can be done with triangulation of police, hospital, and ambulance data~\citep{taylor2016ambulance}. 

\paragraph{Risk of Reoffending Prediction.} \looseness=-1
Criminal justice systems have relied on the prediction of an individual's risk of future crime for more than 200 years. Tools evolved from phrenology's pseudo-science in the 1800s, to actuarial models starting in the 1920-30s, to statistical (e.g., logistic) models dominating for many decades \citep{berk2014using, greene2022forks}. Early examples of `AI' in policing include the HART tool in Durham Constabulary to aid decisions about police bail \citep{oswald2018algorithmic}, assessments of `dangerousness' predicting homicide \citep{berk2009forecasting} and domestic violence \citep{berk2016domestic}. These tools have consistently raised concerns about racial inequity \citep{propublica2016,harcourt2010risk,zilka2023progression}.

\paragraph{Facial Recognition.}
Facial recognition is now used in the UK for both live operations and retrospective investigations~\citep{homeoffice2024facial}. Live facial recognition (LFR) is arguably the most well-known AI application in policing and has been in development for at least 15 years~\citep{davis2011real}. LFR has been scrutinised worldwide since its first pilot, with concerns raised about wrongful arrests and ethnic disproportionality \citep{buolamwini2018gender}. These concerns have been addressed in some instances -- the UK's National Physical Laboratory assessed LFR equity questions \citep{mansfield2023frt}, and the Metropolitan Police have examined false positives, finding that only those on wanted lists were arrested \citep{bbc2025lfr}. However, recent events put these in question (see Box 2).

\begin{table*}[ht]
    \centering
    \renewcommand{\arraystretch}{1.5} 
    \begin{tabularx}{\textwidth}{|X|}
        \hline
        \textbf{USE CASE ASSESSMENT} \\
        \hline
        \textbf{Context:} Use case: \_\_\_\_\_\_\_\_\_\_ \hfill Theme: \_\_\_\_\_\_\_\_\_\_ \hfill Group: \_\_\_\_\_\_\_\_\_\_ \\
        \hline
        \textbf{Risk Classification:} \\
        $\square$ Low risk, can be deployed with minimal checks. \\
        $\square$ Medium risk, deploy only in specific context and with rigorous checks. \\
        $\square$ Unacceptable risk, do not deploy. \\
        \hline
        \textbf{Justification:} \\
        \\
        \hline
        \textbf{Allowed use cases and pre-deployment checks:} \\
        \\
        \hline
        \textbf{Evaluation and monitoring:} \\
        \\
        \hline
    \end{tabularx}
        \caption{Workshop use case feedback form structure.}
    \label{tab:feedback_form}
\end{table*}

\subsection{Racial Bias in Policing AI Systems}

Over the last decade, critical technical and socio-technical research has highlighted bias and related issues in algorithmic tools for policing. Several studies demonstrated that reliance on arrest records as a proxy for underlying crime can cause an exacerbation of racial bias in different types of predictive tools. This is due to the likelihood of a crime becoming known to law enforcement varying significantly with race, sex, and age~\citep{baumer2010reporting, bosick2012reporting, butcher2022_arrests}. \citet{zilka2023progression} demonstrated this for tools predicting the risk of re-offending. Within `hotspot mapping' tools, feedback loops can increase patterns of over-policing in already over-policed communities~\citep{ensign2018runaway, lum2016predict, sankin2021crime}. For facial recognition, early work documented significantly higher error rates for darker-skinned individuals \citep{buolamwini2018gender}, and while technical improvements have narrowed some gaps, equity in practice remains contested and deployment-dependent~\citep{radiya2023sociotechnical}.

\subsection{Participatory Research on AI and Policing}

Alongside the critical work, a growing body of work examines how practitioners and communities relate to data-driven policing. \citet{kearney2024_beyond} interviewed 40 Police Scotland practitioners and found that officers were averse to tools like predictive risk assessment and facial recognition, and were concerned with the erosion of community policing through datafication. \citet{zilka2023exploring} explored police perspectives on algorithmic transparency and found mixed views, highlighting the importance of transparency for public trust alongside operational concerns. Few works have directly engaged community representatives on AI in policing; exceptions include the Citizens' Biometrics Council which deliberated on facial recognition \citep{ada2019beyond}; and \citet{haque2024we} who brought together community members, technical experts, and law enforcement around a crime-mapping application, finding that community members were more likely than domain experts to question the core motivation of the tool rather than its technical execution. \citet{ziosi2024evidence} interviewed community organisations, researchers, and public sector actors about the Chicago crime prediction algorithm, finding that community-impacted groups used evidence of algorithmic bias to centre liberation and healing, while public sector actors used the same evidence to reaffirm existing power structures. Scholars have also raised concerns about algorithmic tools deployed without community input, particularly regarding their impact on marginalised groups~\citep{richardson2019dirty, brayne2017big}.

\subsection{Responsible AI in Policing Frameworks}

Acknowledging that we have moved beyond interest in AI and into live use in policing, several practitioner frameworks for the responsible use of AI in policing have been developed. In the UK, this includes the AI Playbook for Policing~\citep{AIplaybook2025}, the Responsible AI Checklist for Policing~\citep{npcc2025raichecklist}, and the Police Foundation's review of AI in policing~\citep{muir2025}, alongside broader reviews of AI deployment in policing~\citep{zilka2022transparency, berk2021artificial, lee2024effectiveness}. 
At the regulatory level, the landscape is shifting rapidly yet unevenly across jurisdictions. The EU AI Act represents the most comprehensive risk-based AI legislative response to date. It classifies a range of policing AI applications as either prohibited or high-risk. With high-risk deployments subject to mandatory impact assessments, transparency requirements, and human oversight obligations. Critics have noted, however, that the Act's exceptions for law enforcement are broad enough to permit many of the practices it ostensibly restricts~\citep{euaiact2024}.
In the UK, which now,  post-Brexit, sits outside the EU regulatory framework, the approach remains sector-led and primarily voluntary. The Algorithmic Transparency Recording Standard (ATRS), made mandatory for central government departments in December 2024, requires public sector organisations to publish information about how and why they use algorithmic tools. However, policing remains operationally independent: only two police force ATRS records had been published as of 2025, and compliance is not currently mandatory for forces \citep{govuk2024atrs}.


\subsection{The Police Race Action Plan}

UK's Police Race Action Plan (PRAP) \citep{prap2022} was developed jointly by the National Police Chiefs' Council (NPCC) and the College of Policing to address the race disparities affecting Black people that policing cannot currently fully explain. Launched in 2022, the plan sets out key areas for improvement across policing structured around four themes:
\begin{itemize}
\setlength{\leftskip}{-0.5em}
    \item \textbf{Culture and Workforce}: a police service that represents and supports its Black officers, staff and volunteers.
    \item \textbf{Powers and Procedures}: a police service that is fair, respectful, and equitable in its actions towards Black people.
    \item \textbf{Trust and Reconciliation}: a police service that routinely involves Black people in its governance.
    \item \textbf{Safety and Victimisation}: a police service that protects Black people from crime and seeks justice for victims.
\end{itemize}
      The plan's commitments include introducing mandatory training on racism, anti-racism, and Black history; adopting a new ``explain or reform" approach to race disparities in the use of police powers such as stop and search, use of Taser, and other types of force; reviewing misconduct and disciplinary processes to reduce racial disparities; and better enabling Black people to have their voices heard in local communities and within policing itself. The current plan, however, does not include the police's rapid inclusion of new technologies and the associated risk of increased racial bias. This work bridges this gap directly and supplements PRAP with a specific AI-focused mixed-stakeholder deliberation.

\begin{table*}[ht] 
    \centering
    \renewcommand{\arraystretch}{1.4} 
    \begin{tabular}{p{3.06cm} p{9cm} p{3.5cm}}
        \toprule
        \textbf{Theme} & \textbf{Use Case} & \textbf{Risk Classification} \\
        \midrule
        
        \multirow{3}{=}{\textbf{Culture \& Workforce}} 
        & AI in Recruitment & $3, 3, 3, 3, 3$ \\
        &  AI Review of Body-Worn Footage to Identify Misconduct & $2, 3, 3, 3, 3$ \\
        &  Bias-Auditing Dashboards & $1, 1, 3, 3$ \\
        \midrule
        
        \multirow{4}{=}{\textbf{Powers \& Procedures}} 
        &  Predictive Policing (Hotspot Mapping) & $3, 3, 3$ \\
        &  AI-Assisted Classification / Analysis / Summary of Reports & $1, 3, 3$  \\
        &  Risk Assessment / Predictive Profiling & $5, 5, 5, 5$ \\
        &  Live Facial Recognition & $3, 3, 3$ \\
        \midrule
        
        \multirow{3}{=}{\textbf{Trust \& Reconciliation} } 
        &  Community-Sentiment Analysis from Public Data & $1, 1, 1, 1, 3$ \\
        &  Sentiment Analysis / Summary of Community Feedback & $1, 1, 1, 3, 3$ \\
        &  Public-Facing AI Virtual Assistant & $1, 1, 1, 3, 3$  \\
        \midrule
        
        \multirow{3}{=}{\textbf{Safety \& Victimisation} } 
        &  Risk Prediction for Victimisation & $3, 3, 3, 5, 5$ \\
        &  AI Transcription / Translation of Calls / Statements / Interviews & $1, 3, 3, 3, 5$ \\
         & AI Advice to Decide if to Carry the Investigation Forward & $3, 3, 5, 5, 5$ \\
        
        \bottomrule
    \end{tabular}
        \caption{A summary of themes and AI use cases in policing discussed amongst participants. For each use case, the risk classifications are reported from the groups that filled in the use case sheets. Risk levels are enumerated as 1 (low risk or green), 3 (medium risk or amber), and 5 (unacceptable risk or red). A number 2 indicates a group marked between low and medium risk.  }
    \label{tab:themes_usecases}
\end{table*}

\section{Methods}

\subsection{Participants}
Participants were invited to register for the workshop by the organisers based on the recommendations of the central Police Race Action Plan (PRAP) team. Four main groups of participants were invited:  
(a)~community representatives involved in the reduction of racial bias, discrimination, and disproportionality;
(b)~members of the Police working on reducing disproportionality;
(c)~members of the Police working on innovation and AI adoption;
(d)~academics who work on the responsible deployment of AI in policing. 

Individuals were recruited from the PRAP team's networks and rolling recommendations, and---in one case---by a nomination from another invitee. Community representatives who have been or are currently engaged with PRAP were the main source drawn upon for community representation. Several academics were invited by the organisers directly. 
Invitations were sent to  $\approx45$ potential participants, of which 35 registered for the workshop, and 30 attended for the day, excluding 2 organisers. Of those, 15 were police, 11 were community representatives, and 4 were academics. In terms of demographics, approximately half of attendees belonged to UK minority ethnic communities; gender balance was roughly even. 

\subsection{Agenda and Use Cases}

\looseness=-1
After a short introduction, the participants were divided into 6 groups for the morning, and a different set of 6 groups for the afternoon. In each session, the groups reviewed a set of AI use cases relevant to policing, divided into four core PRAP themes: 
(a)~\emph{Culture \& Workforce} (referring to the recruitment, retention and progression of Black officers and staff),
(b)~\emph{Powers \& Procedures} (relating to fair, respectful and equitable use of police powers),
(c)~\emph{Trust \& Reconciliation} (relating to community involvement and anti-racist culture), and 
(d)~\emph{Safety \& Victimisation} (relating to the protection and experiences of Black victims of crime).
Before the workshop, we asked the PRAP team to highlight any use cases they wanted included in the workshop. 
Additional use cases were added by the organisers based on a literature review. The full list of use cases can be found in \Cref{tab:themes_usecases}. 

For each use case, one author created an information sheet that included a deployed example, the intended goals, and known risks associated with the use case. The author attempted to present the information objectively and use non-technical language. The provided information was not comprehensive; it was intended as a conversation starter and leveller, since some participants were less familiar with AI and its associated risks than others. The information sheets for all use cases can be found in the Appendix.

\subsection{Risk Framing and Use Case Feedback}

To structure the discussion, we asked participants to complete an assessment worksheet as shown in \Cref{tab:feedback_form}. As a group, participants were asked to classify the use cases in terms of risk: 
(a)~\emph{Green} -- low risk, can be deployed with minimal checks;
(b)~\emph{Amber} -- medium risk, deploy only in a specific context and with rigorous checks;
(c)~\emph{Red} -- unacceptable risk, do not deploy.
Participants were then asked to justify their selection, indicate allowed use cases and pre-deployment checks, and highlight what needs to be evaluated and monitored. In addition to the workshop, participants had an opportunity to anonymously raise further comments and concerns in a post-workshop feedback survey.

\subsection{Analysis}
After the workshop, the 76 completed worksheets were organised by use case, scanned, and typed by one of the authors. The analysis proceeded in two stages. In the first stage, one author conducted a thematic analysis of the worksheet responses for each use case, identifying recurring concerns, points of consensus and disagreement, and the reasoning participants used to justify their risk classifications. Three additional authors, who were also present in the discussions on the day, reviewed and annotated the emerging themes, ensuring they represented not only the worksheets, but the conversations during the workshop (which were not recorded). We note that the analysis relied on the worksheet responses, and recollections were only used to ensure that there was no misrepresentation of the discussions. In the second stage, two authors examined the questions participants appeared to be implicitly asking when arriving at their risk classifications — what we refer to as back-engineered questions. This was cross-checked by two additional authors to ensure the findings were faithful to the discussions on the day. Where direct quotes are used in the results section, they are drawn verbatim from the worksheets. Risk classification was translated to a 5-point scale, with low being 1, medium 3, and unacceptable risk 5, and summarised for each use case. When a group marked two risk categories (e.g., low and medium), the number in-between the categories (e.g., 2) was assigned. 

\begin{figure*}[tbp]
    \centering
    \begin{boxD}
    \textbf{Individual risk assessment and predictive profiling} tools, such as COMPAS (US), and OGRS and OASys (UK), aim to offer an efficient, consistent, and objective risk categorisation with respect to reoffending. They estimate the risk of re-arrest within a fixed time period (e.g., 12 months). The best models achieve nearly 70\% accuracy. In the UK tools, race is explicitly excluded as an input but validation takes place for different ethnic groups to ensure comparability and minimise inequity \citep{moore2015}. 
    Nonetheless, these tools rely on arrest and/or prior convictions as proxy for previous offending (which tends to be higher than official records). But if the probability of being arrested is itself racially biased, or subsequent justice decisions are, tools will reflect and compound that bias \citep{zilka2023progression}. \\

    This use case gathered the most objections. While two groups did not rank it, the other four groups ranked it as an unacceptable risk. Unlike other use cases where objections are rooted in implementation rather than the idea itself, \textbf{here the objection is clearly to the idea itself} -- ``\emph{bias issues are still baked into this}". Participants also failed to see the practical benefit: ``\emph{would it be [able] to prevent reoffending}?"
    This strong objection is worth noting, particularly as policing rebrands itself as more \textit{predictive} or \textit{precise} as part of its \textit{AI Revolution}. Most tools of this kind focus on re-arrest rather than harm, with no emphasis on rehabilitation, context, or enabling non-reoffending. However, participants acknowledged that the question can be reformulated: rather than predicting who will reoffend, could the question become \textit{what will help this person not to reoffend?} — whether that is housing support, employment, or rehabilitation services. This re-framing is more challenging to implement in practice, but it represents a more constructive and ethically defensible way forward for individualised predictive tools within criminal justice. We note that in the UK, probation risk assessment tools OASys and Asset, include questions on housing, welfare, and benefits with the explicit aim to form part of intervention plans and risk management strategies, which then go on to inform supervision activities \citep{moore2015}.
      \end{boxD}
    \label{fig:placeholder}
    \caption*{Box 1: Strong objection to predicting the risk of re-arrest.}
\end{figure*}

\subsection{Limitations}
Several study limitations should be acknowledged. First, all groups reviewed use cases in the same fixed order, which may have introduced order effects: fatigue, anchoring, or carry-over reasoning from earlier use cases may have shaped how later ones were assessed. Future workshops like this should counterbalance the order of use cases across groups. 
Second, recruitment was conducted primarily through the networks of the PRAP team, which means participants were largely pre-selected for their existing engagement with issues of racial bias and disproportionality in policing. This was a conscious choice as racial bias in policing was central to the workshop; however, we acknowledge this sample is unlikely to be representative of the broader police workforce. One participant in the feedback survey shared that the youth voice was missing. 
This workshop did not aim to present the public's views, but of stakeholders representing diverse interests. 
Third, workshop discussions were not recorded so that participants could feel that their conversations were private and that they could communicate freely. 
Additionally, authors present on the day observed discussions across groups and were able to cross-check that the thematic findings reported were reflective of the conversations as they occurred, and not only of what participants chose to commit to paper. 

Fourth, we acknowledge the possibility of in-group power dynamics. When assigning groups for the workshop, we aimed to balance the police and community representatives and assign one academic to each group. To avoid only dominant voices emerging, we remixed the groups for the afternoon session, such that groups only contained individuals who had not previously been grouped. 
Organisers moved between groups throughout the sessions, encouraging groups to try to allocate even amounts of time to each use case and supporting quieter individuals to join in. Some community representatives appeared to initially feel aware of their lack of AI technical expertise and needed encouragement to share their views \cite{harrington2019deconstructing}. Interestingly, many police present had similar levels of AI technical expertise, but did not seem to need as much encouragement to share. Police participants' confidence in their understanding of how use cases are deployed in practice could translate to confidence talking about responsible AI use. However, the expertise of community representatives regarding ways in which disproportionality is baked into the criminal justice system did not translate as quickly into confidence talking about the responsible AI use. By the end of the day, all participants appeared to feel confident fully engaging and sharing views. Although seven participants completed the post-workshop survey, none of the answers shared additional use-case-related concerns or indications that participants were not able to fully express their views. 



\section{Results of Use Case Discussions}
We now report on the analysis of 76 completed worksheets for 13 AI in policing use cases. Not all 6 groups completed all use cases. Use cases were distributed across morning and afternoon sessions, and groups varied in how many they were able to assess within the time available. In addition, not all completed worksheets assigned a risk categorisation.


\subsection{Theme A: Culture \& Workforce}
For this theme, there were 3 use cases, all marked between low and medium risk by all groups. 

\paragraph{AI in Recruitment.}
The use of automation in recruitment is on the rise in UK policing. Oleeo, a recruiting platform, claims on their website that their platform is used by 70\% of UK police forces~\citep{oleeo2025police}.
The first use case focused on AI in recruitment of new officers, particularly in risk screening.
The main perceived benefits were efficiency and cost savings, and a better experience for the candidates.
Another stated advantage was reduction in bias by hiding the candidate's name and gender, also known as ``\emph{blind recruitment}''.

All groups completed this use case, but only 4 ranked it, assigning medium risk. Participants agreed that diversifying the workforce was mission-critical, but were sceptical of the ability of AI to make headway in this challenge. The existing, non-tech recruitment process was described as ``\emph{a dumpster fire, even before AI}''. Participants found the concept of ``\emph{blind recruitment}'' naive and unlikely to lead to meaningful change, a sentiment backed up by the literature~\citep{simons2021machine}. Resistance to blind fairness was a strong common theme across groups, as was the concern that human-in-the-loop mitigations are proposed without clarity on how they would work in practice. 

Participants  highlighted that training data and assessment criteria reflect existing---non-diverse---norms, which means models can only be trained to identify the types of officers already hired in the past. 
Participants acknowledged the potential benefits of automation, as there may be real gains in time, money, and candidate experience, but highlighted that this might not translate to genuine progress, i.e., to getting the best potential officers or shaping the force to be what it needs to be to future-proof its relationship with a diverse population. Considering the need for change, there was also a worry that automating the hiring process can make it more rigid and harder to reform.
This use case was an example of an overarching theme in the discussions, where participants felt AI is being used as a \textit{shiny fix} for a process with more fundamental problems, without proper evaluations to see whether it brings any real improvements.

\paragraph{AI Review of Body-Worn Camera.}
This use case was relatively well-received, as it adds accountability and does not replace an existing human process. However, participants noted that the underlying assumption---that all behaviour is recorded---is undermined in practice: cameras are sometimes off, a minimum 30-second buffer exists, and officers, particularly bad actors, may use them selectively. This led to scepticism about the system's potential added value. However, most groups agreed there is real potential for improving ``\emph{trust and confidence'', and ``standards and professional behaviour}'', especially if the system should be used not only as a \textit{stick} but also as a \textit{carrot}, with ``\emph{positive behaviour}'' highlighted as well, motivating officers' acceptance and cooperation. 
Fundamental questions were raised about how the act of recording might change the behaviour of everyone involved, and whether it could interfere with helpful, open, and empathetic policing. The surveillance implications for the public depend entirely on how the system is used, what its goals are, and what commitments are upheld. For evaluation, monitoring, ``\emph{community input}'', and ``\emph{dip sampling}'' were suggested, though a concern was raised that supervisors may lack the resources to review footage effectively, increasing the risk of over-reliance on AI.

\paragraph{Bias-Auditing Dashboards.}
Discussions on this use case were briefer than the previous ones, likely due to time constraints. Generally, the prevailing tone was scepticism rooted in a lack of specificity---without understanding how the auditing is done, it is very hard to know whether it can be helpful. Some of that scepticism stemmed from the act of quantification itself and the absence of cultural context: ``\emph{understanding context is more useful than surface-level disproportionality}''. 
Several participants suggested limiting the audit to HR ``\emph{to keep [it] low risk}'' or keeping it project-specific to ``\emph{keep [a] tight scope}''.
Groups who rated this use case as low risk did so on the basis that it is not a decision-making tool but one that offers an opportunity for reflection and transparency, potentially more useful for monitoring AI decision-making rather than human decision-making. Notably, none of the discussions mentioned the potential for ethical whitewashing or other cynical uses of such a tool.

\subsection{Theme B: Powers \& Procedures}

\paragraph{Predictive Policing (Hotspot Mapping).}
Perhaps surprisingly, participants did not object to predictive policing or hotspot mapping per se---the ability to plan and strategise resource deployment was viewed positively. 
Instead, the objections concerned implementation and usage. 
Participants highlighted two core issues: 1) the reliance on arrest data as the primary input, and 2) the goal of optimising for enforcement. 
Arrest data does not faithfully represent harm, victimisation, or community sentiment on feeling unsafe, and crime non-reporting patterns are often systematically biased. 
Participants echoed concerns from literature \citep{ensign2018runaway} about feedback loops, i.e., predictions informing interventions, which generate new data used for further predictions, compounding existing patterns of over-policing of racial minority communities. 
For almost all groups, discussion of data led naturally to discussion of goals, and a clear consensus that increasing arrests is \emph{not in itself a worthy goal}. Participants feared that this technology supports what might be called a \textit{whack-a-mole model} of crime prevention, noting crime displacement and a lack of connection to wider policing strategy or the social causes of crime. One comment captured this well: there is a need to ``\emph{balance between public health and justice approaches}''. On a positive note, participants were optimistic about the potential to use this technology as a resource not just for enforcement but for multi-agency collaboration, strategy, and policymaking, employing wider data sources and a broader range of outputs.

\paragraph{AI-Assisted Processing of Incident Reports.}
\looseness=-1
There is a clear appetite for using AI in this context, provided it is deployed safely and responsibly. The motivations articulated by participants---extracting ``\emph{patterns and trends}'', informing new officers about cases, and managing large volumes of information---all centre on AI as an assistant supporting human decisions, rather than as an automated decision-maker. However, this is often not the guiding principle in the design of current tools. 
For AI-generated summaries and Copilot tools, resistance came from several directions: 
(a)~a false sense of efficiency (``\emph{is it quicker or do you have to spend time fixing the products}'', or worse, it may not be verified at all); 
(b)~accountability questions around whether officers can be held responsible for AI-generated reports without undermining the efficiency gains; 
(c)~lack of contextual, cultural, and compassionate understanding; 
and (d)~potential downstream consequences for prosecution and court procedures. Evaluation is a major, non-trivial effort -- ``\emph{red-team[ing]}'', ``\emph{stress-test[ing]}'', ``\emph{hallucination}'' rates, ``\emph{traceability}'' of data, and ``\emph{information retrieval}'' accuracy are all important, but collectively represent a significant undertaking that may quickly become outdated. At its core, the discussion about risk-benefit boiled down to one central question: where, when, and how is AI adoption worth it?

\paragraph{Risk Assessment (Predictive Profiling) \& Live Facial Recognition.}
Individualised Risk Assessment gathered the most objections of any of the discussed use cases. In contrast, the use of Live Facial Recognition was generally well accepted under certain conditions. Both use cases are discussed in detail in Box 1 and Box 2, respectively.

\begin{figure*}[!tbp]
    \centering
    \begin{boxD}
    \textbf{Live Facial Recognition (LFR)}---as deployed by the Metropolitan Police (the Met; serving the greater London area)---functions as a real-time aid to help officers locate wanted individuals from a watchlist, with safeguarding as an additional declared benefit. Historically, commercial facial recognition systems were shown to have higher rates of misidentification for Black individuals, and Black women in particular \citep{buolamwini2018gender}. However, LFR systems have since improved, and the Met claims that their system can be operated at settings where there is no statistically significant difference in demographic performance across groups \citep{mansfield2023frt}.
    \\

   Within our workshop, LFR was considered acceptable by the majority of groups, but only under clearly defined conditions: 1) use only against ``\emph{high-harm crime}''; 2) do not use for minor crimes (e.g.,``\emph{cannabis use}''); and 3) mitigate racial and ``\emph{intersectional}'' bias, and check for ``\emph{watchlist bias}''. Additional conditions included transparency about who is on the list, where and how the LFR is deployed, and transparency around outcomes and actual bias levels, alongside human oversight, governance, and community consultation.
   The discussion was more refined than for other use cases, likely due to considerable effort by UK policing to communicate the conditions under which LFR is deployed and to demonstrate positive outcomes. The bias discussion extended beyond technical accuracy to include watchlist composition and deployment locations as potential sources of bias. Evaluation criteria included false and true positives, value of use and outcomes, and bias across multiple dimensions. \\

    After the workshop, in March 2026, Essex Police paused LFR deployments after a commissioned study found significantly different performance for Black individuals \citep{bland2026lfr}. Specifically, it found that using a threshold that prevented false identification of Black individuals resulted in under-policing, i.e., the system misses non-Black wanted individuals more often. This meant fewer false arrests, but a disparate impact on Black individuals on the watchlist. 
    In April 2026, the High Court dismissed a judicial review challenge to the Met's revised LFR policy, finding it imposed sufficient constraints to meet the ECHR quality of law test; the court acknowledged that a policy authorising discriminatory use could undermine its legality, but was not persuaded this policy does so. 
    Our findings and these events underscore how acceptance of LFR hinges on specific conditions and assurances about its use and impact.
    \end{boxD}
    
    \label{fig:box2}
    \caption*{Box 2: Live Facial Recognition deemed acceptable for high-harm offenders.}
\end{figure*}

\subsection{Theme C: Trust \& Reconciliation}
\paragraph{Sentiment Analysis \& Summary of Community Feedback.}
These use cases are quite similar, with the former focusing on analysing public data such as social media, and the latter on feedback submitted to the police. Both use cases were generally perceived as low to medium risk. The key justification was that the scale of monitoring involved is ``\emph{not possible to do manually}''. Therefore, AI analysis can function as an informative layer on top of existing processes rather than replacing them. Described as a useful ``\emph{temperature check}" and tool for emerging issues, sentiment analysis was considered acceptable when used to inform rather than directly drive action.
The risks identified were consistent across both use cases: the sentiment captured is not a balanced picture and can be skewed by ``\emph{keyboard warriors and loud voices}"; the tools could exclude ``\emph{some community members}'', e.g., offline people and ``\emph{marginalised voices}''; and the tools may struggle with nuance, sarcasm, tone, and local community context. Evaluation is also a challenge---outputs require comparison to outputs from ``\emph{other methods of gathering public opinion of the police}'', e.g., human assessment; also, evaluations should not just be focused on efficiency. The recommendations across groups were to ``\emph{explore other data analysis techniques before deployment}'', and to remain clearly aware of the data's limitations.

\paragraph{Public-Facing AI Virtual Assistant.}
Classified as low to medium risk, the virtual assistant was considered acceptable when limited to ``\emph{non-emergency}'', low-risk, and informative purposes. One group noted that the ``\emph{public expect[s] these type of tools}'', and already uses external LLMs to ask questions about information on police websites. A key distinction was drawn between a chatbot that simply directs users to the right part of a website (low risk but also low benefit), and one that engages more substantively, which raises greater concerns around misinformation, safeguarding, and public trust.
Essential requirements included:  monitoring for effectiveness, ``\emph{dwell times and response outcomes}'', and user feedback; a redirection-to-an-operator mechanism; and clear signalling about appropriate use (e.g., do not use in an emergency; do not report a crime through this channel). The concern that ``\emph{misinformation can reduce public confidence}'' was raised, alongside a cost-benefit question: ``\emph{how many people end up calling 101 anyway?}'' Accessibility and the ability to detect vulnerability in user queries, something a human operator might more naturally recognise, were also flagged as important design considerations.

\subsection{Theme D: Safety \& Victimisation}

\paragraph{Risk Prediction Tools for Victimisation} See Box 3. 

\paragraph{AI Transcription / Translation of Victim Calls / Statements / Interviews.} 
This use case did not achieve consensus, being rated low risk by one group but unacceptable by another, with several medium risk assessments in between. The group that rated it low focused on compliance with existing regulations (e.g., CPIA\footnote{The Criminal Procedure and Investigations Act, 1996.}), data security, governance, and officer training -- a procedural approach. The group that rated it unacceptable was chiefly concerned with the risk of victims and witnesses being misunderstood, and its potential disparate impact: ``\emph{negatively affect the Black community due to cultural misunderstandings}". This concern about disproportionate impact was shared broadly among the participants, who observed that ``\emph{minority groups are already a step behind, so AI increases the discrimination}''.
Concerns were raised about where and how transcription would be used, with non-evidential recordings considered more acceptable, and higher-stakes uses, such as evidence presented at court or 999 calls, considered not acceptable. 
Most groups were sceptical that automated transcription would result in genuine time savings, if done properly: ``\emph{how much time is being saved if everything is being checked over?}" Requirements included human verification of output, transparency, consent, and appropriate control for the victim or witness, including the ability to ``\emph{stop, start, and [go] back}". Questions of accountability were also raised: ``\emph{who takes ownership of ensuring that [the transcript] is correct?}"

\paragraph{AI Advice to Decide Whether to Carry an Investigation Forward.}
Three of five groups classified this use case as unacceptable risk, with the remainder classifying it as medium risk. The most accepting groups saw potential value if used as ``\emph{part of a suite of products to support decision-making}", with the tool providing assistance rather than acting as a \textit{decision-maker} in itself. The groups who found it unacceptable focused on the risk of disparate impact and bias, as well as concerns about gamification and broader ethical concerns: ``\emph{seems completely unethical}" and a defiance of ``\emph{bare minimum of what policing should be doing}''. Even the more accepting groups expressed concerns about bias, over-reliance, the effect on victims, and whether the tool would deliver true efficiency compared to a human baseline.

\begin{figure*}[!tbp]
    \centering
    \begin{boxD}
    \textbf{Risk prediction tools for victimisation}---such as Spain's VioGén \citep{alvarez2018integral,satariano2024viogen}---aim to reduce harm by predicting risk in domestic violence cases. However, their accuracy is fundamentally limited, not by technology, but the difficulty of the prediction problem. 
    A particular blind spot is the lack of recording of interventions, which makes cases where violence was successfully prevented incorrectly classified as low risk.
    In the UK, the checklist-based risk-assessment tool, DASH, was designed to help frontline practitioners identify individuals at highest risk of serious harm, but a series of tragic failures raised concerns that it was missing high-risk victim-survivors, failing to capture dynamic risk, the lived realities of victim-survivors, the cultural nuance affecting minority communities, and has little use in terms of predictive validity \citep{turner2019dashing}.
    In 2022, the College of Policing recommended switching to DARA, a tool which has better predictive validity \citep{messing2013average}; yet as of August 2025, 20 of 39 UK police forces still use DASH. \\

    Our groups' views were divided: two rated this use case as unacceptable risk, three as medium risk, and one left it unmarked. Where deemed acceptable, justifications included the tool already being in use in some form, the potential for AI to enable greater scale and consistency, and its value as a ``\emph{second voice}'' that may reduce human bias. 
    It was emphasised that the tool should function as ``\emph{part of a wider toolset / factors for consideration}'' to support officer decision-making, as it was ``\emph{high risk if used as a sole tool without giving consideration of other factors}''.  
    The subtlety of this assessment was stressed throughout, with cases ``\emph{often requir[ing a] nuanced understanding of context to make [a] holistic decision}''. 
    A key requirement was that officers retain full accountability ``\emph{to remain accountable for decisions made}''. This was tied directly to bias (``\emph{fear of overreliance, lack of accountability on bias from police}''), and an overall worry that the tool would exacerbate cultural stereotypes: ``\emph{Black women are seen as less vulnerable so large bias in training data}''. The importance of participatory development was highlighted: ``\emph{bring in victims/survivors - validating questions being asked, gather their perception on this}'', alongside the need for multi-agency response. 
    For evaluation, the need to understand the human baseline was emphasised. However, this is inherently difficult, as successful interventions prevent harm and may therefore resemble misjudgments of risk, making it hard to demonstrate good judgment precisely when it succeeds.
    \end{boxD}
    \label{fig:box3}
    \caption*{Box 3: Concerns around Risk prediction tools for victimisation exacerbating racial bias and cultural stereotypes.}
\end{figure*}

\section{Analysis of Risk-Bounding Process}

In addition to considerations on the specific use cases, the workshop helped us gain insights into the \emph{process} of establishing risk categories. Participants were asked to classify the use cases; they received no guidance on what to consider, except that it should be in the context of racial bias. 
Below, we examine the reasoning process that emerged. 

Our main finding is that although the explicit framing was racial bias, the reasoning process that emerged was considerably broader. Rather than focusing exclusively on risks to racial minorities, participants consistently reasoned about whether a tool delivers genuine benefit for everyone, including, but not limited to, marginalised communities. The questions groups asked were therefore not only about who might be harmed, but about whether the tool works, for whom, under what conditions, and with what safeguards. We examine this reasoning process in detail below, and return to its broader implications in the discussion.

\paragraph{Does It Work?} The starting point for most groups was whether the tool is actually capable of doing what it claims. In some cases, this was a matter of the technological capability or the availability of good-quality, non-biased data. However, for several use cases---including recidivism and victimisation risk prediction---the objections were fundamental and concerned whether the task itself is tractable: predicting victimisation risk, for instance, is a hard prediction problem regardless of the sophistication of the tool. The question of whether the tool works was therefore not only technical but conceptual -- does the use case rest on a sound premise?
A sub-question was ``\emph{will it effectively help a human decision maker?}'' As participants had strong objections to autonomous decision-making, groups consistently questioned whether the tool is enabling the officer to make better judgments and decisions, or effectively deciding for them how to act. This distinction was central to how groups assigned risk: tools perceived as adding an additional layer of information were more accepted, while those perceived at risk of displacing human judgment received higher risk ratings. Participants deemed it crucial that accountability for outcomes and for bias can be meaningfully upheld.

\paragraph{Will It Deliver Genuine Benefit?}
Even for uses deemed potentially acceptable, groups pressed hard on whether deployment would translate into meaningful benefit in practice. One concern was the false sense of efficiency: groups asked whether time savings are real if AI is used responsibly, i.e., when appropriate verification and correction are accounted for, and whether the tool will actually save money if it opens the organisation up to legal challenges. More importantly, participants questioned whether the tools can deliver benefits beyond efficiency: ``\emph{Will this use of AI help alleviate the current structural problems or worsen them?}''
Some groups had a more pragmatic approach to considering the risk-benefit analysis, using the flawed status quo as a reference point. Groups considered, given the current process being problematic or already under-resourced, whether AI is likely to help. This was often emphasised in the context of bias: if the officers are biased, can ``\emph{giving them AI reduce overall bias, or will it make it worse?}''

\paragraph{Will It Deliver Benefit for \emph{Everyone?}} \looseness=-1
The third and most consistently applied layer of scrutiny asked whether the benefits of a tool would be distributed equitably, and specifically, whether marginalised communities would share in them, or bear a disproportionate share of the risks.
Some of these discussions were rooted in well-known failure points such as biased data, but much of the discussion focused on how loss of context and nuance around language and culture can lead to disparate impact. 
Crucially, these questions were framed not only in terms of harm avoidance but also of benefit: would the tool help the force become more equitable? Would it improve outcomes for victims from minority communities? Would it build or erode trust?
This inclusive framing meant that racial bias was not treated as a separate checklist item but was woven into the broader assessment of whether the tool delivers on its promises---for everyone, not just those already well-served by existing processes.
Questions of community involvement followed naturally: groups asked who defines what \emph{good} means, whether community input has been sought, and whether the change is likely to improve trust and relationships with those most affected by policing.

\paragraph{Measuring Success and Public Benefit.}
Alongside these questions was a practical one: can we actually measure or estimate whether a tool works, and whether it delivers benefits equitably? This requires, as a starting point, a clear and measurable definition of ``\emph{what does success look like?}'', and an understanding of ``\emph{what evidence is needed to say that it's effective?}'' A strong condition for acceptance was that use cases be designed to have demonstrable benefits, beyond a simplistic notion of efficiency, and that these are actively monitored, measured, and communicated transparently.

\section{Discussion}
This is a critical time for policing, which faces strong competing pressures: the desire and public expectation to deliver a better, more protective service; the urgent imperative to address deep and longstanding racial inequity; and the reality of shrinking resources. AI is frequently positioned as the solution to this tension -- a way to do more with less, and a clear win when done responsibly~\cite{AIplaybook2025, moore2025concerning}. In this framing, efficiency is the primary promise, and concerns around ethics, including racial bias, are treated as blockers to unlocking enormous benefits. The reality, however, is more complex: there are very few, if any, clear examples of AI in policing delivering genuine, measurable public benefit, yet numerous examples of algorithmic tools exacerbating already unacceptable racial disparities \citep{propublica2016,ensign2018runaway,sankin2021crime,zilka2023progression}. This paper presents evidence that challenges this framing, showing that considering racial bias is not an obstacle but a critical lens to be applied as early as possible when considering AI adoption.

\subsection{The Importance of Community Consultation}

\looseness=-1
Although many concerns raised by participants align with findings from responsible AI literature, the risk categorisations did not always follow expected lines. For example, hotspot mapping---a use case with well-documented equity concerns \cite{ensign2018runaway, lum2016predict}---was relatively well received; participants viewed the underlying goal of strategic resource deployment as legitimate and the problems as rooted in implementation rather than the idea itself. Similarly, LFR was generally deemed acceptable, despite considerable attention to racial bias concerns~\citep{buolamwini2018gender, radiya2023sociotechnical}. In contrast, recidivism risk assessment, deployed for decades in the UK and worldwide, gathered the strongest objections of any use case discussed, critiquing the idea itself, not only its technical execution. Participants questioned if such tools could deliver benefit to the public, or help with offender rehabilitation.
Indeed, these tools were introduced to \emph{reduce} racial bias and redirect low-risk offenders from prison, but evidence shows they have failed on both counts~\citep{kleinberg2018human, zilka2023progression, stevenson2024algorithmic}. This distinction---between objections to implementation vs to the premise---is hard to surface by any means other than community engagement, and it reiterates its importance. 
The workshop also surfaced many cultural and contextual concerns rooted in lived experience, such as adultification and the perception of Black women as less vulnerable.

\subsection{Racial Equality as a Key to Better Design}
Current approaches to responsible AI in policing tend to rely on structured, qualitative evaluation frameworks -- responsible AI checklists, impact assessments, and long-form documentation that works through considerations of bias, transparency, accountability, and other principles one by one (e.g., \citep{npcc2025raichecklist, gds2025dataethics}). These frameworks serve an important function, but they implicitly treat ethical considerations as a separate line of enquiry rather than a fundamental lens for deciding whether a use case is worth pursuing at all. The deliberative benefit-risk process we observed in this workshop was strikingly different from this checklist model. 
Participants did not work through a list of considerations; instead, they reasoned in an integrated way, with many of these considerations flowing naturally towards a set of fundamental questions: does it work? Will it deliver genuine benefit and will that benefit extend to everyone?
We found this question-led deliberation process much more reflective of participatory design~\cite{jackson2023participatory, labedzka2026missing}, with the strong underlying vision that a tool that can deliver benefits to marginalised communities will also benefit the police and the public as a whole. This vision mirrors what is known as the curb-cut effect \citep{blackwell2017curb}: designing with marginalised users in mind surfaces better questions, and is likely to lead to better outcomes for everyone. We argue that for AI in policing, best practice should be not just designing with marginalised users in mind, but, as much as possible, with them in the room \citep{hamraie2017building}.

\subsection{People Before Progress}

\looseness=-1
Our findings demonstrate that discussion of AI use cases need not be all-or-nothing~\cite{Lawal2026}. Participants were clearly not against technological progress or AI: they ranked only 3 out of 13 use cases with an average risk above medium. 
In fact, participants in this mixed-stakeholder workshop were more accepting of AI use cases than police professionals alone in a similar study~\citep{kearney2024_beyond}. Despite much discussion of human bias and flawed current processes, participants maintained a clear stance in favour of human judgment, consistently distinguishing between tools that inform human judgment and tools that risk displacing it~\cite{elish2019moral,gao2021human}. 
Indeed, designing AI to support rather than supplant human judgment may be the most effective way to realise its benefits while managing the associated risks.
Recent work on AI for missing persons investigations illustrates what this may look like in practice: a hybrid system combining LLM-based summarisation with rule-based visualisations and source linking, designed to augment officers' sensemaking while maintaining autonomy and professional accountability~\cite{labedzka2026missing}. This shows that AI can be built to genuinely help officers---and, through them, the communities they police---rather than creating a veneer of objectivity while quietly eroding the space for professional judgment.


\section{Conclusion}
This work demonstrates that mixed-stakeholder deliberation, centred on racial equity, is both feasible and valuable at the earliest stages of AI adoption in policing. Community representatives brought knowledge that technical review alone cannot replicate -- surfacing concerns rooted in lived experience and adding invaluable context. Foregrounding racial bias did not narrow or politicise the deliberation; it deepened it. This reinforces the notion that ethical considerations and participatory approaches should be at the centre of AI adoption — particularly when better public service, not just efficiency, is the true goal.

\section*{Acknowledgments}

MJ was supported by the Engineering and Physical Sciences Research Council [Grant Number EP/Y009800/1 AI KP0003] through funding from Responsible Ai UK. AS's time was funded as part of his independent contractor role for the Met Police. MZ acknowledges support from the Leverhulme Trust grant ECF-2021-429, which also covered the costs of the workshop itself. We are very grateful to Abena Akuffo-Kelly, Ashwin Varghese, Dave Hudson, Detective Chief Superintendent Shaun White, Dr Colin Burton, Dr Doirean Wilson, Israr Hussain, Jonathan Vale, Judith Germain, Khi Rafe, Laura Toop, Lizzie Rapinet, Megan Julien-Harris, Michael Maher, Paulette Johnson-Clarke, Sam Darbyshire, Professor Myrtle Emmanuel, and Professor Robert Binns for dedicating their time and sharing their views for this work. 

\section*{Ethical Considerations Statement}

This work involved human participants via an in-person workshop. Attendees took part in a professional or representative capacity rather than as private individuals; data were recorded via group worksheets and conversations were not recorded. All quotations are reported without attribution, with demographic information given only in aggregate. During the workshop, groups were balanced between police and community representatives, with groups remixed between sessions. All participants were provided an anonymous post-workshop feedback channel.

\paragraph{Representation.} This workshop does not attempt to represent or reflect the general public opinion, and it should not be read as doing so. Community representatives were recruited through networks connected to the Police Race Action Plan, and most had existing engagement with policing institutions. This was deliberate as it produced participants with deep, specific knowledge of racial disproportionality within and related to policing. However, it also means our participants were people willing to work with the police towards reform. Communities most harmed by policing include those who regard engagement itself as the wrong strategy, and may reject the premise that these tools should be assessed rather than opposed. Those views are absent here, and although this was not deliberate, it is likely a consequence of our recruitment strategy. A workshop constituted differently may have produced different classifications. However, our findings describe many relevant and useful concerns and considerations and the reasoning behind the choice of risk categorisation. We emphasise that our results do not necessarily reflect the range of community opinion, and our participants' acceptances of AI use cases that we report should be read with that in mind.

\paragraph{National context.} This work was conducted in the UK, where policing operates under the notion of \textit{policing by consent} and where the vast majority of officers do not routinely carry firearms. Both conditions shape what participants treated as proportionate. Judgements about the acceptable risk, for example of a misidentification, inevitably rest on assumptions about how an encounter is likely to escalate. In jurisdictions with routinely armed policing, different foundations for police legitimacy, or different levels of state violence, the same use case will carry different stakes. Our deliberative method is portable; however, the specific risk classifications (Table 2) are not, and should not be transferred across jurisdictions.

\subsection*{Adverse Impact Statement}

\textbf{This paper does not endorse any policing technology.} We report how a group of stakeholders reasoned about a set of use cases. We do not claim that the use cases they found acceptable are acceptable, that they are safe or effective, or that they should be adopted. No classification in this paper constitutes a recommendation. Our finding that participants were broadly open to most use cases, if quoted in isolation, could be presented as community endorsement. However, from the findings it is clear that even when use cases were classified as low risk, they were not endorsed and many conditions and caveats were applied to every case. Detaching a classification from those conditions will misrepresent the findings in this paper. We would regard it as a misuse of this work for any police force, vendor, or policymaker to cite it as support for adopting a technology, or as evidence that communities have been consulted about one.

Our contribution is to the question of how these decisions should be made and who should be present when they are, as well as important conditions and concerns.

\subsection*{Positionality Statement}
The authorship team approaches this work primarily from an academic perspective, with expertise in responsible AI and algorithmic fairness. One author brings a more practitioner-oriented experience, and another has a background in public policy. All authors believe that racial inequality is a critical issue in policing and the criminal justice system, and this may have shaped the analysis. One author's time was funded through a contractor role with a police force, which we note as a potential conflict of interest. Although members of the PRAP team helped with recruitment and identifying use cases, no police force or funder had editorial control over this paper, reviewed it prior to submission, or had any right of approval over its findings.

\bibliography{bib}

\includepdf[pages=-]{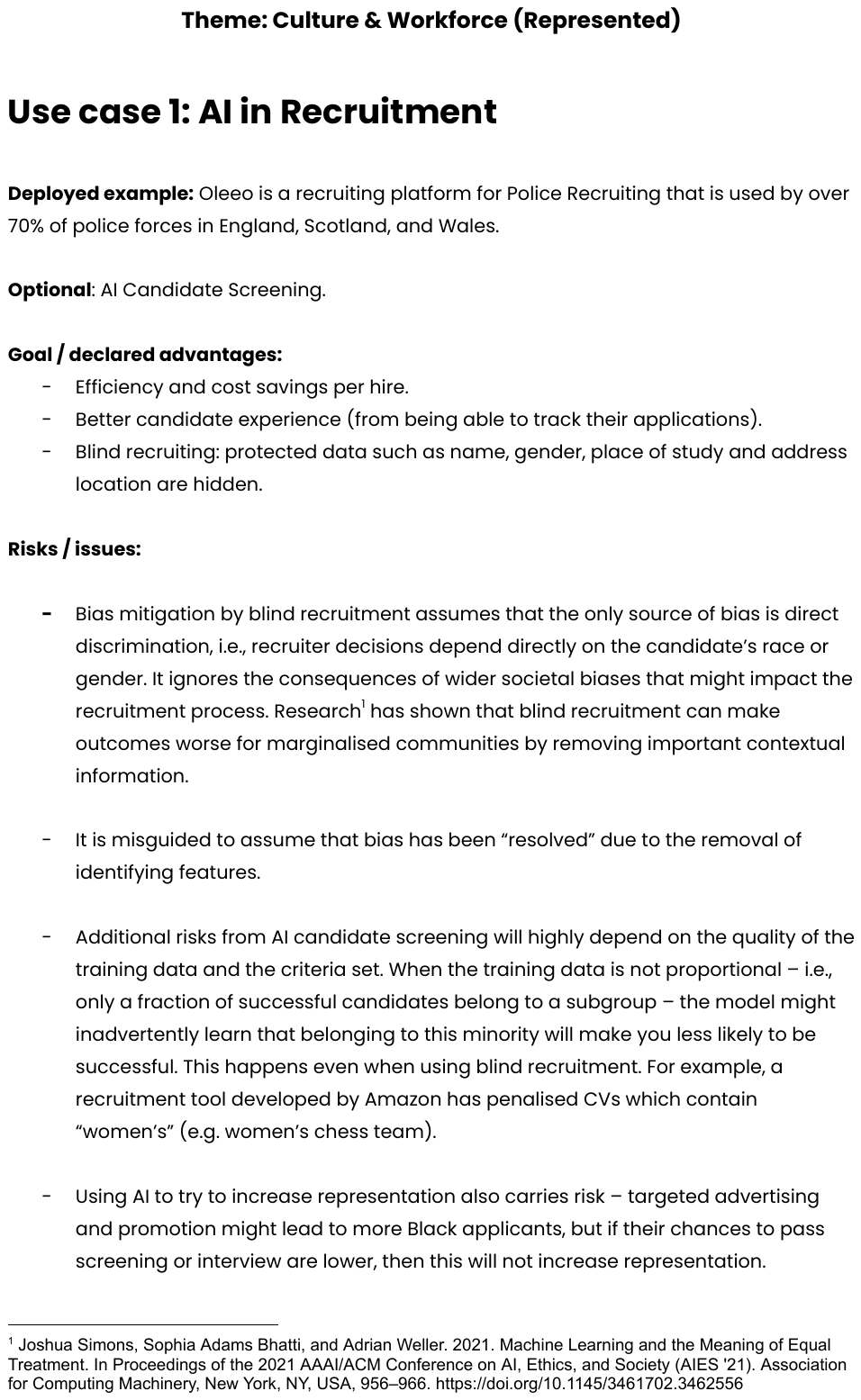}   
\includepdf[pages=-]{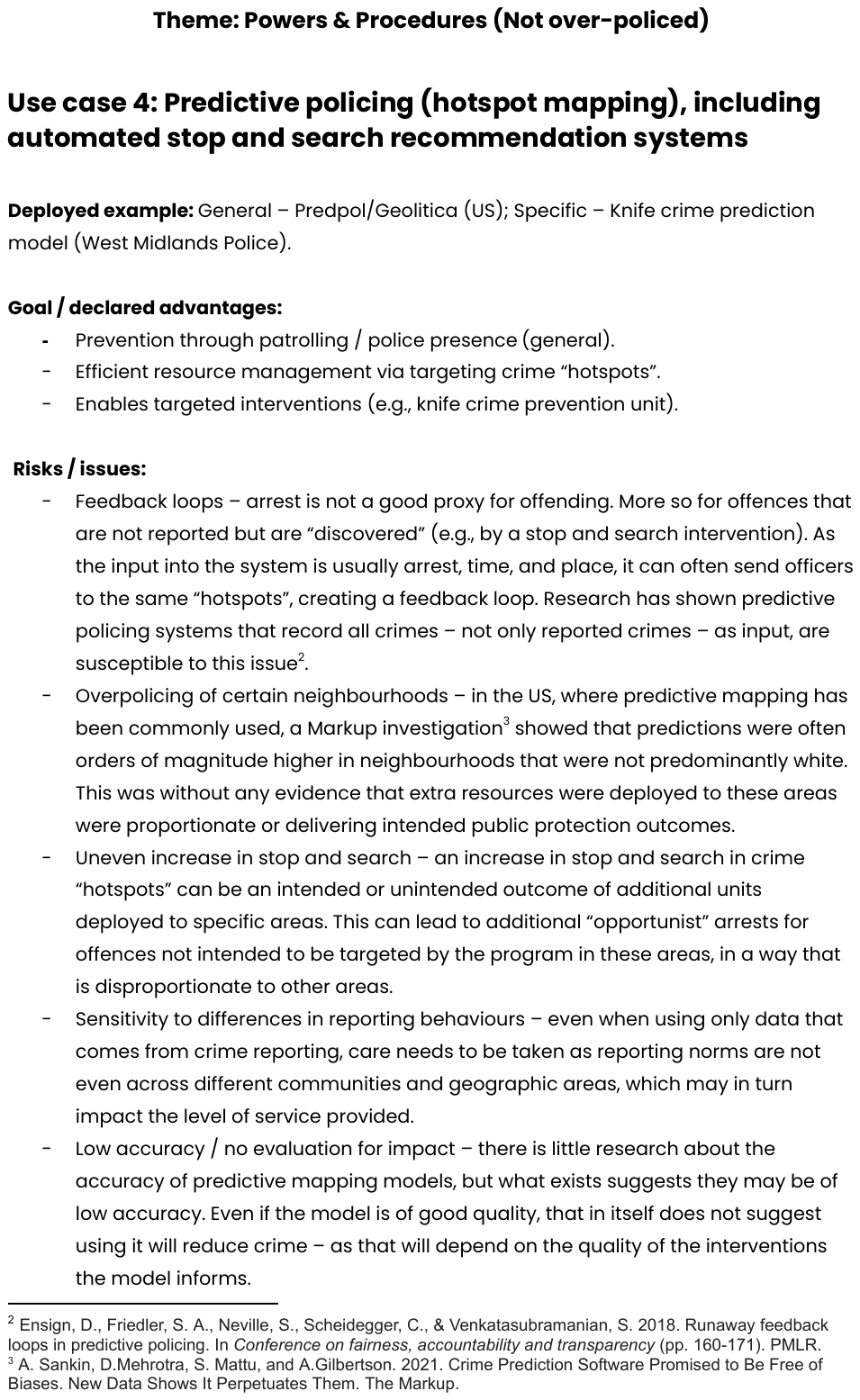}   
\includepdf[pages=-]{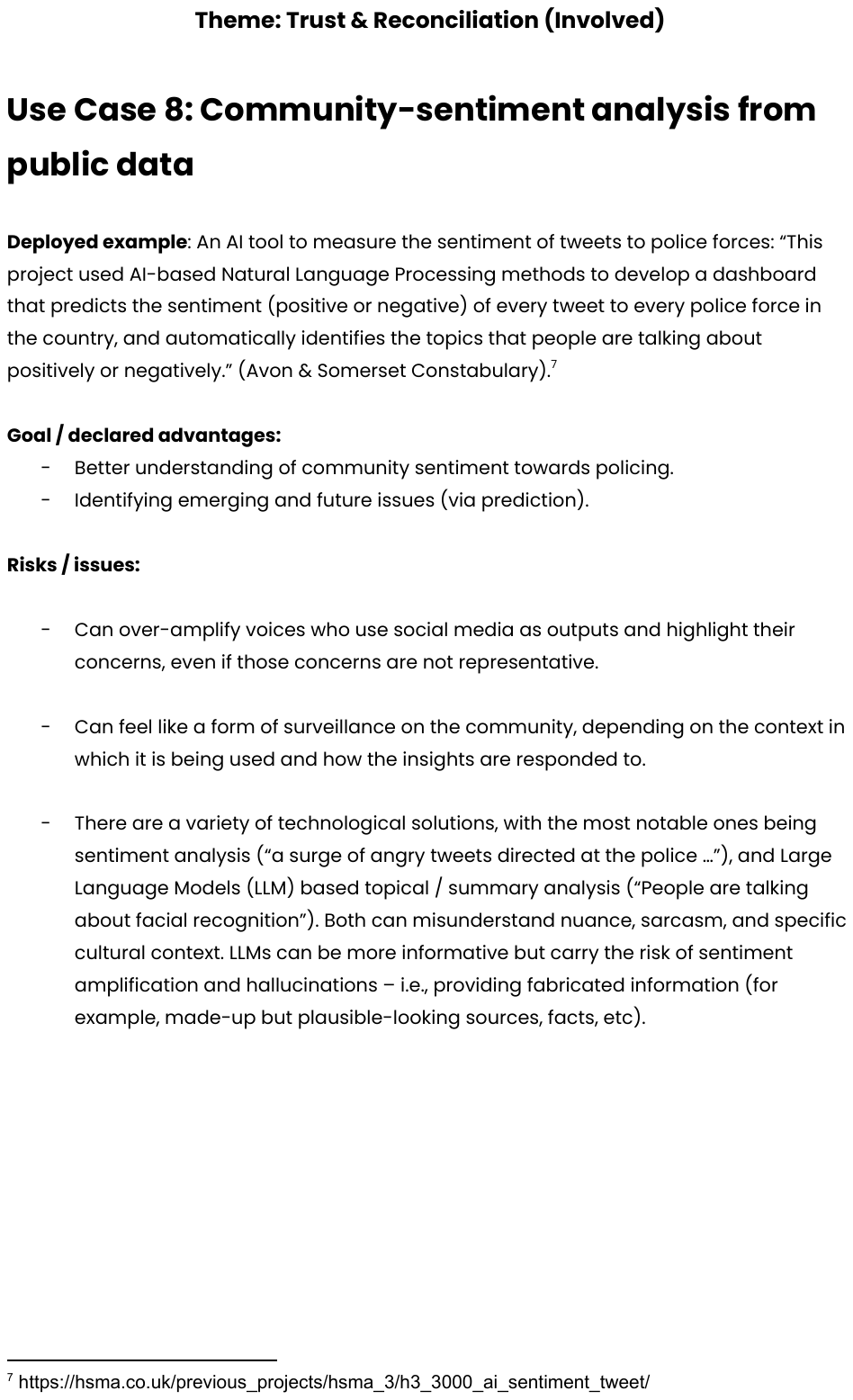} 
\includepdf[pages=-]{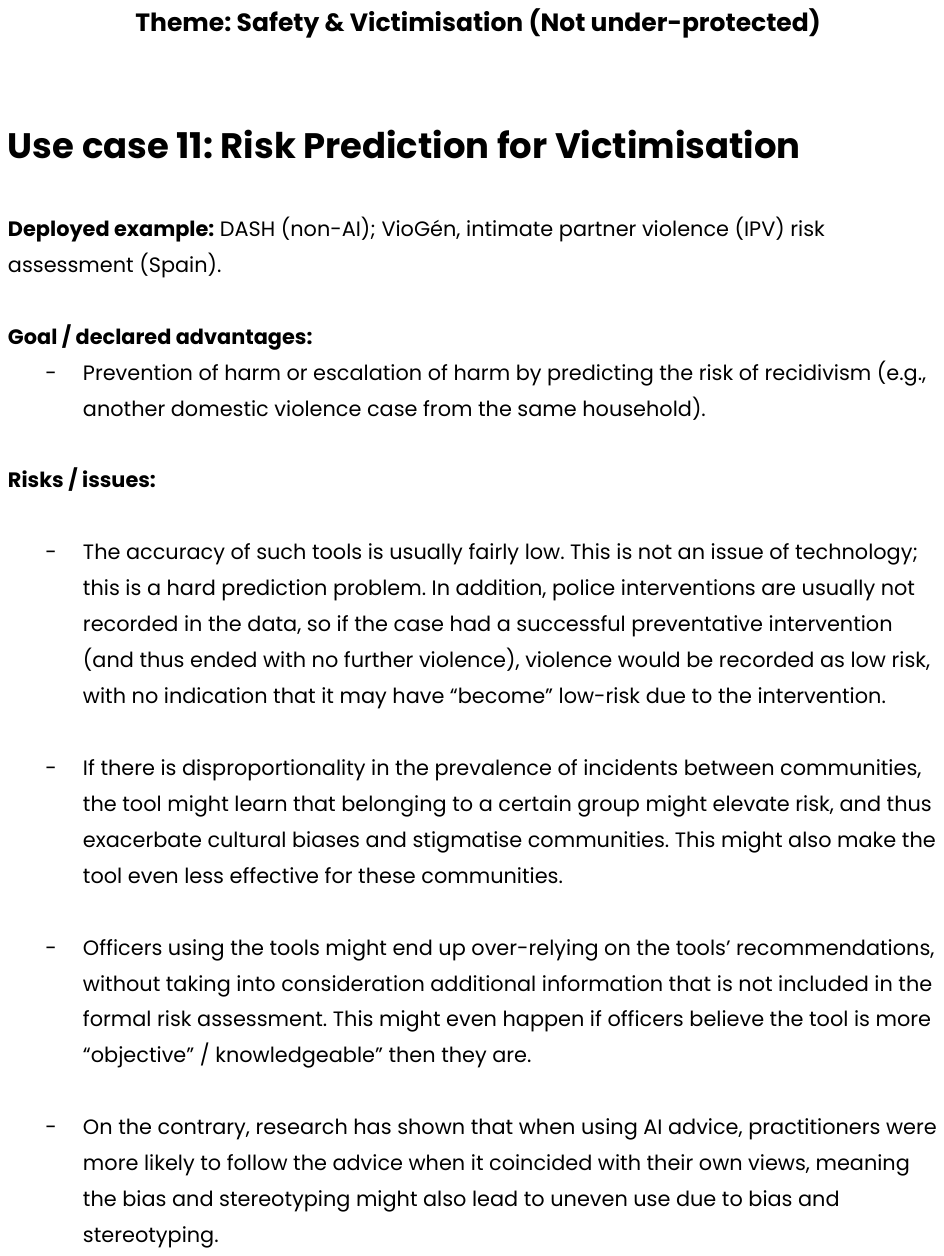}   

\end{document}